\documentclass[conference]{IEEEtran}

\IEEEoverridecommandlockouts
\usepackage{times}

\usepackage[numbers]{natbib}
\usepackage{multicol}
\usepackage{amsmath}
\usepackage{graphics}
\usepackage{epsfig}
\usepackage{amssymb}
\usepackage{subfig}
\usepackage[english]{babel}
\usepackage{algorithm}
\usepackage{algorithmic}
\usepackage{url}

\long\def\MET#1{{\bf ---#1---}}
\long\def\SS#1{{\it ---#1---}}
\long\def\MET#1{{}}
\long\def\SS#1{{}}

\DeclareMathOperator*{\R}{\mathbb{R}}

\begin{document}

\title{Autonomous Waypoint Generation with Safety Guarantees: On-Line Motion Planning in Unknown Environments}

\author{\IEEEauthorblockN{Sanjeev Sharma}
\IEEEauthorblockA{Graduate Student of Computing Science\\University of Alberta, Canada, Email: sanjeev1@ualberta.ca}}
\IEEEpubid{Author did this work in 2013, when he was a graduate student in the Computing Science department at the University of Alberta}
\maketitle

\begin{abstract}
On-line motion planning in unknown environments is a challenging problem as it requires (i) ensuring collision avoidance and (ii) minimizing the motion time, while continuously predicting where to go next. Previous approaches to on-line motion planning assume that a rough map of the environment is available, thereby simplifying the problem. This paper presents a reactive on-line motion planner, Robust Autonomous Waypoint generation (RAW), for mobile robots navigating in unknown and unstructured environments. RAW generates a locally maximal ellipsoid around the robot, using semi-definite programming, such that the surrounding obstacles lie outside the ellipsoid. A reinforcement learning agent then generates a local waypoint in the robot's field of view, inside the ellipsoid. The robot navigates to the waypoint and the process iterates until it reaches the goal. By following the waypoints the robot navigates through a sequence of overlapping ellipsoids, and avoids collision. Robot's safety is guaranteed theoretically and the claims are validated through rigorous numerical experiments in four different experimental setups.  Near-optimality is shown empirically by comparing RAW trajectories with the global optimal trajectories.
\end{abstract}

\IEEEpeerreviewmaketitle

\section{Introduction}
\label{Introduction}
Autonomous vehicles and robots will soon help, or perhaps replace, humans in the tasks that are  
time consuming, hard or unsafe. Such tasks include autonomous driving, search and rescue, patrolling and engaging with enemy forces --- often confronting unseen, hostile and unstructured environments. A common challenge underlying these tasks is efficient on-line motion planning: deciding where to go next, ensuring collision avoidance and minimizing the motion time. Thus, the trajectory generator for such vehicles and robots should be able to (i) minimize the motion time, (ii) decide where to go, and (iii) ensure collision avoidance.

This paper presents an on-line reactive motion planner, Robust Autonomous Waypoint generation (RAW), for mobile robots navigating in unknown and unstructured environments. RAW generates a locally maximal ellipsoid, using semi-definite programming, around the robot, separating it from the surrounding obstacles. A reinforcement learning agent then makes a decision about where to go and generates a waypoint inside the ellipsoid. The robot then navigates, remaining inside the ellipsoid, towards the waypoint. By following the waypoints to the goal, the robot navigates through a sequence of (overlapping) ellipsoids --- a virtual, obstacle free, tunnel. RAW assumes that the robot has limited field of view and has access to its coordinates  or relative position to the goal.

On-line motion planning for mobile robots has been a focus of the robotics community for the past three decades. The previous approaches can be broadly categorized as: (i) assuming that a complete map of the environment is available; (ii) using some known structure in the environment; and (iii) assuming that a global path or a global guidance function is available (as may be derived from a potential function or through a graph search). Fox {\it et al.}~\cite{DWindow} introduced dynamic window to search for admissible velocities in a given time frame, to avoid collisions in known environments. Fiorini and Shiller~\cite{ShillerVO} proposed velocity obstacles for motion planning in known environments. Velocity obstacle based approaches require carefully tuning the time-step parameter to avoid poor performance. 
Shiller {\it et al.}~\cite{ouriros,IJRR1} introduced an on-line planner that generates a sequence of intermediate goals in the environment, and produces a trajectory to the goal, passing through the intermediate goals. However, the approach assumes that a full map of the environment is available. Sampling based approaches, like RRT~\cite{lavalle}, have been successfully demonstrated in many robotics applications. However, RRT based methods are in general limited to known environments.

Quinlan and Khatib~\cite{ElasticBand} proposed elastic band for optimizing an initial collision free configuration-space path to the goal. The path is adjusted (and smoothed) on-line for sensor based obstacle avoidance. However, it assumes that the deviations from the global path are small. If the deviations are large, the approach may fail, requiring the global path to be recomputed using the updated information (also see~\cite{GeometricallyUnknownEnvironments} and~\cite{ElasticStrips} for similar works). The assumption of availability of such a global path to the goal is invalid in this paper. \IEEEpubidadjcol

Local path-set approaches were used by many finalists of the DARPA Urban Challenge, successfully demonstrating its application to motion planning in unknown environments. Kuwata {\it et al.}~\cite{MIT} presented non-fixed random path-sets using RRT. A similar work is fixed path-set, presented by Knepper~{\it et al.}~\cite{InformedPathSampling} (and references therein). Path-set methods rely on a global guidance function to guide the search towards the goal, and often constructing such a function requires map of the environment. In the DARPA Urban Challenge, the structure of the problem (following a road) was used as a heuristic to constrain the paths to end parallel to the road. Such a heuristic, or a navigation function, may not be available in unstructured and unknown environments. Minguez and Montano~\cite{EgoKinoDynamicSpace} introduced the ego-kinodynamic space (E-KS), building a local space for generating trajectories and reactively avoiding collisions. E-KS takes the robot's stopping distance into account and guarantees safety within a certain time interval. However, (i) it uses potential fields to guide the robot to the goal, and potential fields assume that a map of the environment is available; and (ii) with limited field of view, the safety may not be guaranteed at future time-steps --- a trajectory along which the braking constraint is not yet active may lead to collision with an obstacle not observed in the current view.  A similar approach is anytime re-planning using a graph search in the local high-dimensional lattice-space~\cite{TimeBoundedLattice,LocalGlobalLikhachev}. The robot executes a global path obtained using the graph-search and re-planning is done upon receiving the updated information. The changes in the environment are assumed to be small, or the global plan may fail, thereby requiring the global-search to be re-performed.
\section{The Proposed Algorithm: RAW}
Having discussed the underlying simplifications of the previous approaches to on-line motion planning, this section now presents the proposed approach for on-line motion planning in unknown and unstructured environments. The previous work~\cite{AWGS} presented an autonomous waypoint generation strategy (AWGS) for on-line path planning in unknown and unstructured environments. In AWGS, a reinforcement learning~\cite{RL98} (RL) agent analyzes the local surroundings of the robot and then generates a waypoint in its field of view (FOV). The robot then navigates to the waypoint and the process iterates until it reaches the goal. AWGS builds a novel representation that makes the RL agent's policy environment independent --- the policy is learned in one environment and can be reused in novel environments without requiring relearning. This makes AWGS suitable for mobile robot navigation in unknown environments. However, AWGS assumes that the robot can execute arbitrary motion commands to avoid collisions, and therefore disregards the safety concern. A robot with kinodynamic constraints, such as a minimum turn-radius constraint, cannot move arbitrarily to avoid collisions. Thus, the robot's safety is not guaranteed in AWGS. To ensure safety of the robot, a waypoint generated at current time-step, say $t_{current}$, must ensure safety at any time $t>t_{current}$. 

In AWGS, the agent learns its policy using a reward function that encourages it to reach the goal quickly, while penalizing for collisions with the obstacles. Designing a reward function that balances safety and optimality is a challenging problem~\cite{Undurti, Undurti2}. A too large penalty for collisions may make the RL agent too conservative and it may never navigate through a tight space, even if a collision free path exists. A too low penalty may result in collisions as the RL agent will try to reach the goal quickly to get a higher reward, ignoring the collisions. Furthermore, when faced with robot's motion constraints, the RL agent may fail to recognize a waypoint that may result in a collision in future. To ensure the robot's safety, the problem can be formulated as a constrained Markov decision process~\cite{Undurti2} (C-MDP). However, solving a C-MDP is computationally very expensive, requiring a solution to a mixed-integer linear program or an exhaustive look-ahead search with branch and bound, both limited to a small finite (discrete) state-space and require full map of the environment.

\begin{figure}[t]
\begin{center}
\includegraphics[trim =2mm 4mm 10mm -1mm, width=6cm]{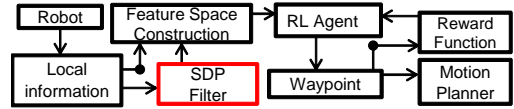}
\end{center}
\captionsetup{font=footnotesize}
\caption{RAW uses semi-definite program to filter out potentially dangerous waypoint (SDP filter) locations, to guarantee robot's safety in unknown environments. Modification to AWGS is indicated using a red box.}
\label{fig:CRP}
\vspace{-.7cm}
\end{figure}

RAW addresses the safety issue by forming a locally maximal ellipsoid, using semi-definite programming (SDP), around the robot and discards the waypoint locations that lie outside the ellipsoid. The surrounding obstacles are constrained to lie outside the ellipsoid. Any two consecutive ellipsoids overlap with each other such that the robot lies in the region of intersection of the two ellipsoids. Thus, if the robot's initial configuration is feasible, i.e. separated from the surrounding obstacles by an ellipsoid, then the robot's trajectory is guaranteed to be collision free. As shown in Figure~\ref{fig:CRP}, RAW uses AWGS architecture for waypoint generation, but removes the potentially dangerous waypoint locations (SDP Filter). The ellipsoid effectively filters out the infeasible actions using SDP.

The contribution of this paper is an on-line reactive motion planner for mobile robot navigation in unknown and unstructured environments that (i) guarantees collision avoidance with unforeseeable obstacles, assuming that the robot's initial configuration is feasible, and (ii) produces trajectories that are not far from the optimal trajectories.  In RAW, the RL agent takes a locally-feasible optimal action at each planning cycle. However, in unknown environments, no theoretical guarantee can be provided on the path optimality. Thus, RAW trajectories are compared with the optimal trajectories and those generated by the RRT, to show that the sequence of locally optimal actions generate reasonably good trajectories (when the environment has no dead-ends, as discussed later). The next section discusses AWGS; section-\ref{MotionPlan} describes the robot model and waypoint generation with robot's motion constraints; section-\ref{SDP} describes the SDP for filtering the waypoints to ensure the robot's safety; section-\ref{Experiments} discusses the empirical results; and section-\ref{Conclusion} concludes the paper.
%
%
%
%
%
%
\section{Notation, Assumptions and Background}
\label{background}
It is assumed that: (i) the robot's field of view (FOV) is limited; (ii) obstacles in the FOV are static and fully characterized; and (iii) the robot has access to its coordinates $z_t$, at time-step $t$, and the coordinates of the goal $z_\textit{g}$. The robot's configuration (position and orientation) at time $t$ is represented as $C_t$. Obstacles in the FOV are represented using a point cloud, $O_t$, of $m$ obstacles at time $t$; $v^T$ represents the transpose of a vector $v$; and $S^2_{++}$ represents the space of all $2\times 2$ positive definite symmetric matrices.
%
%
%
%
\subsection{Reinforcement Learning: Markov Decision Process (MDP)}
\label{RLSec}
In an MDP, a state-action value function $Q^\pi(s_t,a_t)$ for a policy $\pi$ is the expected return of taking action $a_t$ in state $s_t$ and then following $\pi$. The probability of taking an action $a_t$ in $s_t$ is $\pi(s_t,a_t)$. The RL agent's task is to learn $\pi$ that maximizes the expected sum of discounted future rewards from any $s_t$, with discount factor $\gamma\in[0,1)$. By taking action $a_t$ in $s_t$, the agent makes a transition to state $s_{t+1}$, and receives a reward $r_t(s_t,a_t,s_{t+1})$. $Q^\pi$ is approximated using a linear function approximation architecture: $Q(s_t,a_t)=\langle\phi(s_t,a_t),{w}\rangle$, where $\phi(s_t,a_t)\in\mathbb{R}^k$ is the state-action feature vector for $(s_t,a_t)$ and $w\in\mathbb{R}^k$ is learned using the samples.
\begin{figure}[t]
\begin{center}\vspace{0.1cm}
\subfloat[]{\label{fig:gridpoints}\includegraphics[trim =30mm 35mm 60mm 28mm, clip, width=2.5cm]{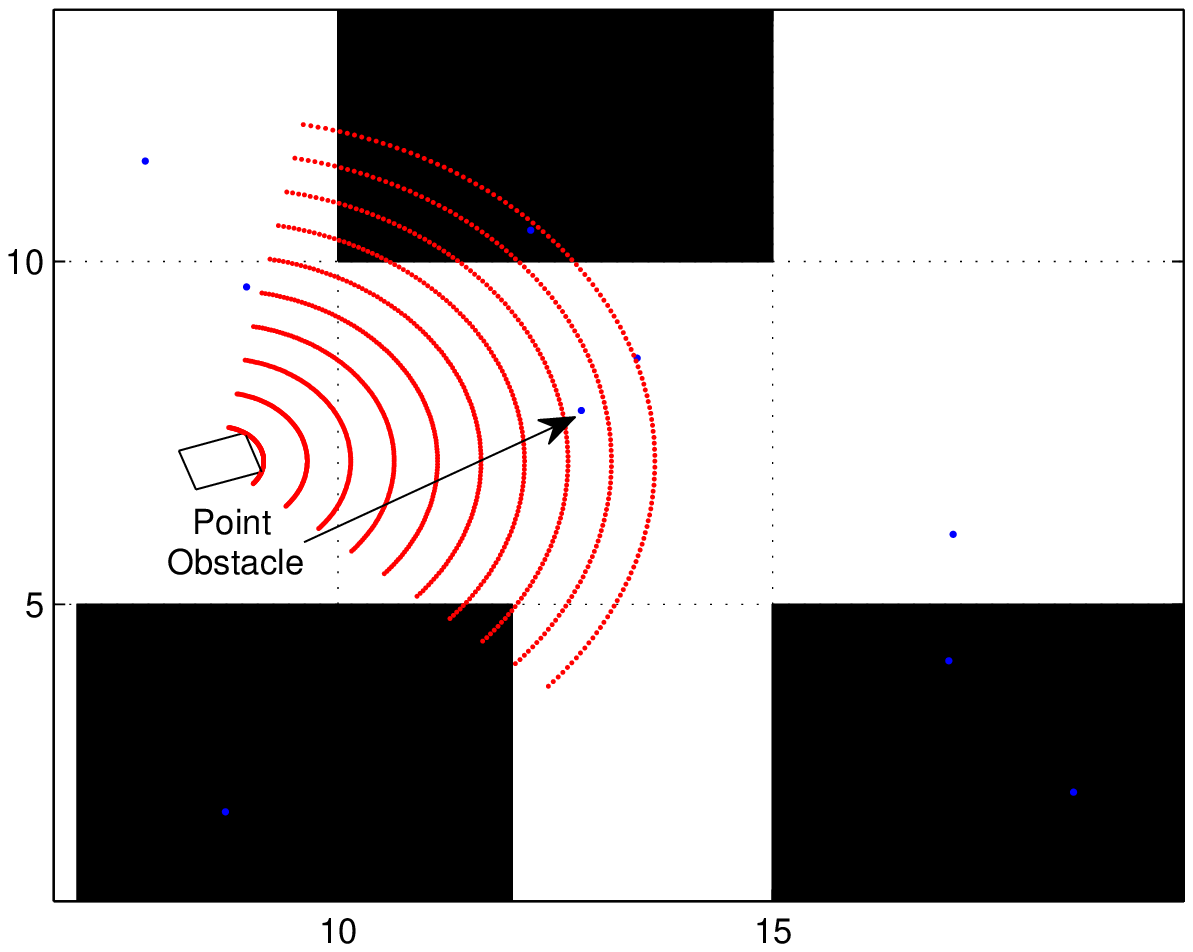}}\text{  }\text{   }
\subfloat[]{\label{fig:PM}\includegraphics[trim =30mm 35mm 60mm 28mm, clip, width=2.5cm]{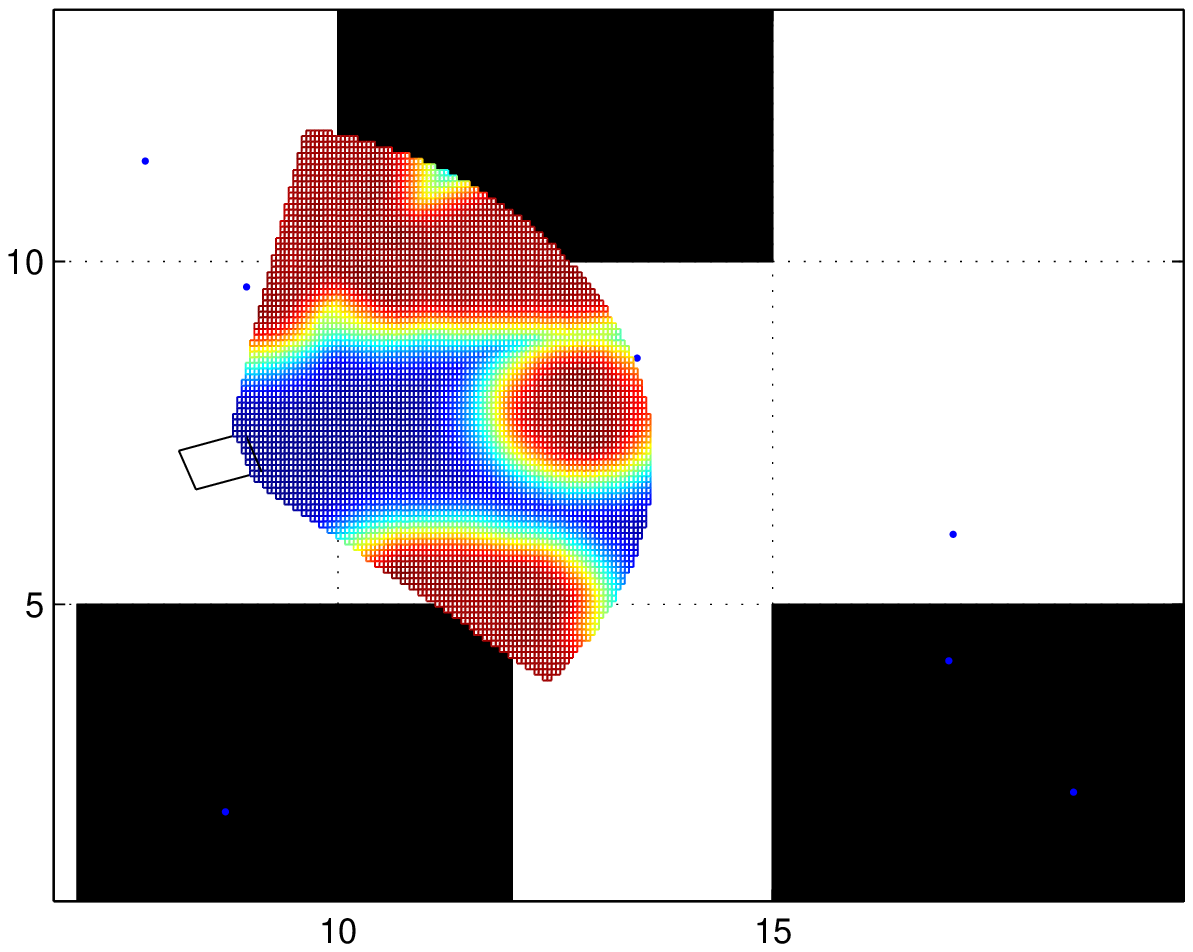}}
\end{center}\vspace{-.4cm}
\captionsetup{font=footnotesize}
\caption{(a) shows the grid-points as red-dots and (b) shows the potential map; red regions represent higher (i.e., dangerous) grid-point values.}
\label{fig:FOVgrid}\vspace{-.5cm}
\end{figure}
%
%
%
%
\subsection{Autonomous Waypoint Generation Strategy (AWGS)}
\label{AWGS}
In AWGS, an RL agent analyzes the robot's FOV. The local region in the FOV is represented by a potential map, which requires descretizing the FOV. The FOV is defined by $r\in(0,R_\textit{FOV}]$ and $\theta\in[-\theta_\textit{FOV},\theta_\textit{FOV}]$; $\{R_\textit{FOV},\theta_\textit{FOV}\}$ control the robot's view. $r$ and $\theta$ are descretized in steps of $dr$ and $d\theta$. The total number of discrete grid-points, $N$, in the robot's FOV is $((2\theta_\textit{FOV}/d\theta)+1)R_\textit{FOV}/dr$. One of these grid-points is then selected as a waypoint by the RL agent. The obstacles in the FOV are represented as a point cloud $O_t$, at time-step $t$.  A potential map $V\in[0,1]^N$ is then computed;  the $i^\textit{th}$ element of $V$ is the potential of the $i^\textit{th}$ grid-point. Figure~\ref{fig:FOVgrid} shows the grid-points and corresponding potential map.

Next, a feature space is constructed for the RL agent for generating a waypoint. The feature space is constructed using the potential map and three geometric parameters for each of the grid points in the FOV. The first two geometric parameters for a grid-point compute the progress towards the goal if that grid-point is selected as the waypoint. The third parameter $\zeta_j=1$ if a straight line path to the $j^{\it th}$ grid-point collides with an obstacle, and $\zeta_j=0$ otherwise. 
The robot's current position represents the agent's state, while an action corresponds to selecting one of the grid-points as the waypoint. Thus, at each time-step, the RL agent has $N$ possible actions. Once the waypoint is generated, the robot follows a local trajectory to the waypoint for time $\Delta T$. The RL agent learns a policy using a reward function that penalizes the agent for generating the waypoints in obstructed regions of the FOV (when $\zeta_j=1$) or close to the boundary of obstacles (as measured by grid-point's value in potential map). For selecting the $j^\textit{th}$ grid-point as a waypoint, it receives a reward:
\begin{equation}
\label{RW1}
reward = -10^3\zeta_j -\alpha_1V_j + \max\{\alpha_2{500},-5\},\nonumber
\end{equation}
where $\alpha_1$ is a constant that controls penalty for defining the waypoints close to obstacles, and $\alpha_2=1$ if the waypoint is defined at the goal and is $-1$ otherwise. $\max\{\cdot\}$ returns either $+500$ or $-5$, encouraging the agent to reach the goal quickly.  
%
%
%
%
\section{Motion Planning: Car-Like Robot}
\label{MotionPlan}
This paper considers a simple (Reeds and Shepp's) car-like robot shown in Figure~\ref{fig:Car}. The configuration of the robot in space at time $t$ is $C_t = (x_t,y_t,\theta_t)$ --- its position ($z_t$) and orientation (with respect to $x$-axis) at time $t$. The distance between the front and real axle, $L$, is $1\,m$ and the minimum turning radius is also $1\,m$. The coordinates of four corners on the robot's body (a rectangle), according to its current configuration $C_t$, are represented using a matrix $\Xi(C_t)\in \R^{4\times2}$. The kinematic model of this robot is:
\begin{equation}
\label{RSModel}
\begin{pmatrix}
\dot{x},& \dot{y}, &\dot{\theta} 
\end{pmatrix} = \begin{pmatrix}
{\cos \theta},& {\sin \theta},&0 
\end{pmatrix}u + \begin{pmatrix}0,& 0,&1 
\end{pmatrix}v,
\end{equation}
where $u\in\{-1,+1\}$ describes the linear velocity and $v\in [-1,+1]$ describes the angular velocity. The magnitude of linear velocity is always $1\,m/s$. Given any two configurations $(x_1,y_1,\theta_1)$ and $(x_2,y_2,\theta_2)$ in space, a time-optimal trajectory connecting them can be computed efficiently~\cite{WangCDC}.
%
%
%
%
\subsection{Safety Issues: Waypoint Generation}
\label{IC}
\begin{figure}
\begin{center}\vspace{-0.0cm}
\subfloat[]{\label{fig:Car}\includegraphics[trim =15mm 8mm 15mm 2mm, clip, width=4.2cm]{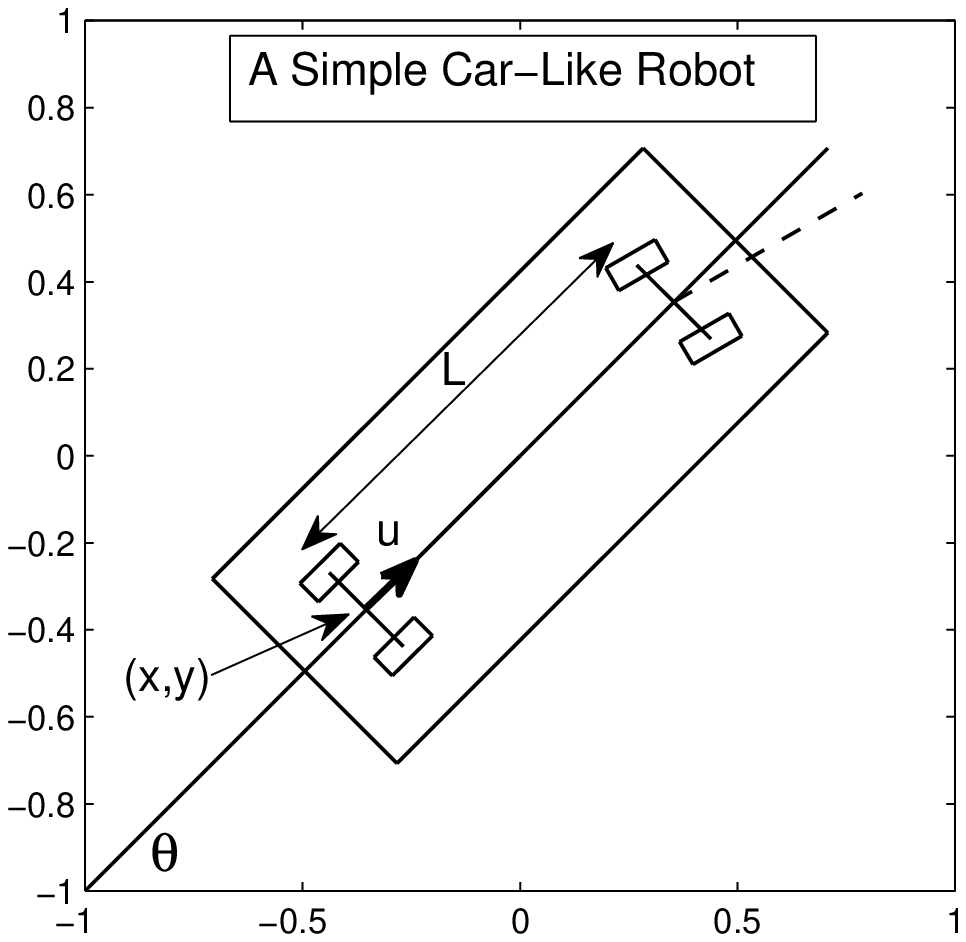}}\,\,
\subfloat[]{\label{fig:Block}\includegraphics[trim =14mm 10mm 12mm 8mm, width=4cm]{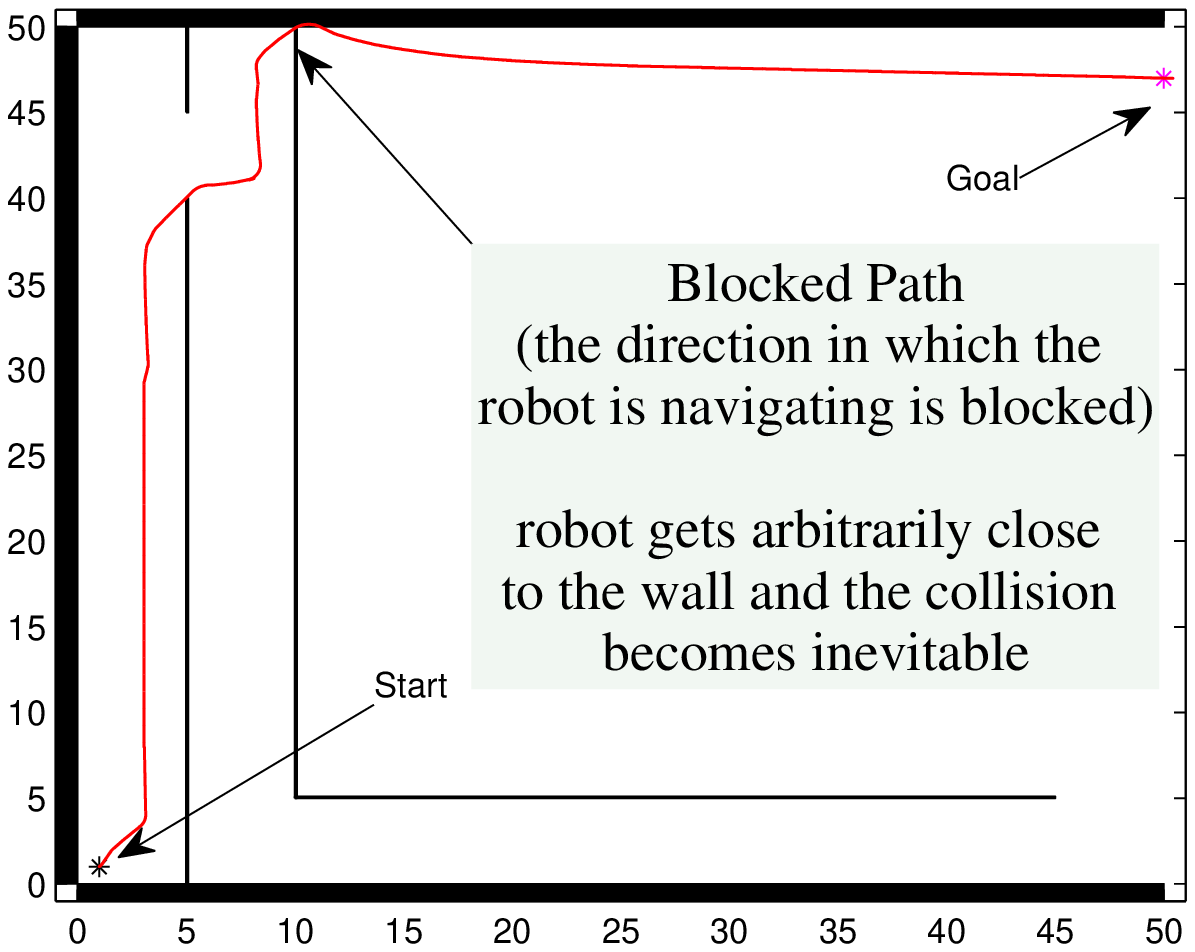}}
\end{center}
\captionsetup{font=footnotesize}\vspace{-0.4cm}
\caption{(a) A simple car like robot; and (b) shows a situation where the robot gets close to the wall and crosses it to reach the goal at the top-right corner, i.e. the safety of the robot is not guaranteed in AWGS.}
\vspace{-0.5cm}
\end{figure}

This section addresses the challenges faced by AWGS with the robot's motion constraints in unknown environments. Figure~\ref{fig:Block} shows that the robot is navigating through a sequence of corridors to reach the goal at (50,50). However, due to limited visibility, as the robot comes too close to the corner at around (10,50), the collision becomes inevitable. The agent did not take the unforeseeable obstacles in to account, and generated waypoints that led the robot too close to the obstacles. As all the possible waypoint locations result in collision, the RL agent generates a waypoint that crosses the obstacles. The waypoint generation process thus does not account for possible future collisions --- disregarding safety. 
RAW addresses this issue by filtering out the potentially dangerous waypoint locations, thereby limiting the actions available to the RL agent, using SDP. The RL agent thus selects waypoints that guarantee collision avoidance at future time-steps.
\subsection{Convex Semi-Definite Programming (SDP): Filtering}
\label{SDP}
This section discusses the SDP for ellipsoid generation, which filters out the potentially dangerous waypoint locations. The ellipsoid is generated such that the robot lies inside the ellipsoid, obstacles lie outside and the goal may lie inside the ellipsoid. Also, as discussed earlier, the RL agent selects one of the $N$ grid-points in the FOV as a waypoint. Thus, the ellipsoid tries to incorporate as many grid-points as possible, inside it. Let the robot's configuration be $C_t$. Let $\Xi(C_t)^k, k = \{1,...,4\}$ represent $(x,y)$ coordinates of the $k^\textit{th}$ corner on the robot's body. The ellipsoid $\Psi_t$, parametrized by $P_t\in S^2_{++}, q_t\in \R^2, r_t\in \R$, at time $t$ is represented as $\Psi_t = \{x\in \R^2|x^TP_tx + q_t^Tx + r_t\leq0\}$. Let $z_{i}$ be the location of $i^{\textit{th}}, i=\{1,...,m\}$, point obstacle in the cloud $O_t$. Let $\gamma_j\in \R^2$, $j=\{1,...,N\}$, be the location of $j^\textit{th}$ grid-point in the robot's FOV. Let $|P_t|$ denote the determinant of $P_t$. The SDP for ellipsoid formation and filtering is:
\begin{eqnarray}
\label{MSP}
&&\min_{P_t,q_t,r_t,\lambda,\nu}\,\,\,\,\nu + \log(|P_t|^{-1}) + \sum_{j=1}^N\lambda_j \,\,\,\,\,\,\textbf{s.t.}\,\,\,\, P_t\succeq I; \nonumber\\
&&\Xi(C_t)^kP_t(\Xi(C_t)^k)^T + \Xi(C_t)^kq_t + r_t \leq -1\nonumber\\
&&z_i^TP_tz_i + q_t^Tz_i + r_t \geq 1, i=\{1,...,m\}\nonumber\\
&&\gamma_j^TP_t\gamma_j + q_t^T\gamma_j + r_t\leq -1 + \lambda_j, j=\{1,...,N\}\nonumber\\
&&z_g^TP_tz_g + q_t^Tz_g + r_t \leq -1 + \nu;\,\,\,\, k=\{1,...,4\}\nonumber.
\end{eqnarray}
$\lambda_j$ and $\nu$ are the slack variables, allowing the soft constraints for including the $j^\textit{th}$ grid-point and the goal, respectively, inside the ellipsoid. If $\lambda_j\geq1$, then the $j^\textit{th}$ grid-point is marked as an infeasible waypoint location. The constraint $P_t\succeq I$, where $I\in S^2_{++}$ is an identity matrix, is a positive definite constraint on matrix $P$. The objective $\log(|P_t|^{-1})$ minimizes the volume of the ellipsoid. This keeps the volume of ellipsoid in check, as the inclusion of slack variables ($\lambda_j$ and $\nu$) in the objective results in an expansion (as much as possible) of the ellipsoid (constrained by the surrounding obstacles). The hard constraints for the robot and the surrounding obstacles separate the robot from the obstacles. Minimizing $\nu$ is equivalent to minimizing the distance between the goal and the boundary of the ellipsoid. Minimizing $\sum_j\lambda_j$ results in an inclusion of as many grid-points inside the ellipsoid as possible. The next section theoretically guarantees the robot's safety.
\subsection{Safety Guarantee}
\label{Safety}
The proof of the safety of the robot follows from the sequence of lemmas and a theorem in this section.

\textbf{DEFINITION:} The region of intersection, $\Gamma_t^{t+1}$, between two ellipsoids $\Psi_t(P_t,q_t,r_t)$ and $\Psi_{t+1}(P_{t+1},q_{t+1},r_{t+1})$ formed at time $t$ and $t+1$, respectively, is: $\Gamma_{t}^{t+1} = \{x\in \R^2|x^TP_tx + q_t^Tx + r_t\leq0, x^TP_{t+1}x + q_{t+1}^Tx + r_{t+1}\leq0\}$.

\textbf{LEMMA-1:} Let $C_t$ be the robot's configuration at time $t$, such that $\Psi_t$ separates $\Xi(C_t)$ from the surrounding obstacle cloud $O_t$. The robot's new configuration $C_{t+1}$ (and hence $\Xi(C_{t+1})$), after it follows the trajectory to the waypoint for time $\Delta T$, remains inside $\Psi_t$.

\begin{proof}The waypoint is always inside the ellipsoid $\Psi_t$. The trajectory to the waypoint, however, may have some segment outside $\Psi_t$. 
However, the robot follows the trajectory only for time $\Delta T$, and there always exists $\Delta T>0$ such that the robot remains inside $\Psi_t$ even when some segment of the trajectory is outside $\Psi_t$. In RAW, auto-tuning of (maximum limit)\footnote{This may be computed analytically for convex-shaped robots, or using a quadratic program for general shaped robots and is discussed elsewhere~\cite{QCQPt}.}
$\Delta T$ is provided by $\Psi_t$ through-out navigation.\end{proof}

\textbf{LEMMA-2} Assume that the SPD is feasible at time $t$ and returns $\Psi_t$. There exists $\Delta T > 0$ such that the robot remains inside $\Psi_t$ during navigation for time $\Delta T$ along the trajectory.

\begin{proof}If $\Delta T=0$, then the robot will have to stop immediately, otherwise it will navigate outside the current ellipsoid, violating Lemma-1. To show that RAW guarantees collision avoidance with unforeseeable obstacles, it is essential to show that the robot always remains inside the current ellipsoid. Thus, it suffices to show that there always exists $\Delta T>0$ such that $C_{t+1}$ and $\Xi(C_{t+1})$ are inside $\Psi_t$. 

Let there be a function $F:\R\times \R^3\rightarrow \R^3$ such that $C_{t+1} = F(\Delta T, C_t)$, i.e. the robot's new configuration after it navigates from its configuration $C_t$ for time $\Delta T$ is given by some (non-linear) mapping $F$. Thus, to prove that there exists $\Delta T\neq 0$, it suffices to show that: $
%
\lim_{\Delta T\rightarrow0^+} \Xi(F(\Delta T, C_t))\in \Psi_t
%
$, where $0^+$ indicates positive side of $0$. This implies that in the worst case, the robot can take infinitesimally small steps and remain inside $\Psi_t$. 
Consider a case where $\Psi_t$ is such that the robot's body touches the the $\{-1\}$ level-set\footnote{An $\alpha$ level-set of $f:\R^n\rightarrow \R$ is defined as: $\{z\in \R^n\,\,|\,\, f(z)\leq\alpha\}$.} of $\Psi_t$. Note that the SDP involves constraints:
\begin{equation}\Xi(C_t)^kP_t(\Xi(C_t)^k)^T + \Xi(C_t)^kq_t + r_t\leq-1, k=1,...,4\nonumber.
\end{equation}
Thus the robot's body can touch $\{-1\}$ level-set of $\Psi_t$. Assume that the robot's trajectory requires it to cross the $\{-1\}$ level-set (for example, moving outside $\Psi_t$). The robot can move until one of the corners of its body (which is a rectangle) touches the $\{0\}$ level-set of $\Psi_t$. Thus, a non-zero $\Delta T$ implies non-zero distance between the $\{0\}$ and $\{-1\}$ level-sets. It therefore suffices to show that the distance between the $0$ and $-1$ level-sets of $\Psi_t$ is non-zero. If this distance is zero, then $\Delta T$ will be $0$ --- the robot already touches the $-1$ level set and therefore cannot move further along any trajectory requiring it to move towards the higher level-sets.  Proving that the distance between the level-sets is non-zero is tedious in the non-canonical form. Thus, the ellipsoid is transformed in to its canonical position using some Euclidean transformation $T_E$ (rotation and translation). Furthermore, with some affine transformation, $T_A$, the canonical ellipsoid can be transformed in to a circle. It then suffices to show that: (A) first and foremost, the resulting circle ($0$ level-set) has non-zero radius $\mathcal{R}_1$; and (B) the radius of the circle corresponding to $-1$ level-set is $\{\mathcal{R}_2\,|\,\mathcal{R}_2\geq0, \mathcal{R}_1-\mathcal{R}_2>0\}$. If $\mathcal{R}_1=0$, then the $\{0\}$ level-set shrinks to a point. Thus, $\{-1\}$ level-set will also shrink to a point --- implying $\mathcal{R}_1-\mathcal{R}_2 = 0$. To avoid clutter, the time-stamp $t$ is removed --- $\Psi_t$ is represented as $\Psi$ and is parameterized by $P\in S^2_{++}, q\in \R^2, r\in \R$. The boundary of the ellipsoid is given by ($x\in \R^2$):
\begin{eqnarray}
\label{bound}x^TPx + q^Tx + r &=& 0, 
\end{eqnarray}
where $P$ and $q$ are of the form:
\begin{eqnarray}
\label{MatP}P = \begin{bmatrix} u_{11}& u_2\\ u_2 & u_{22}\end{bmatrix}; \,\,\,\,\, q = \begin{bmatrix} b_1 \\b_2\end{bmatrix}; u_{11}, u_{22}>0; u_2, b_1,b_2\in \R\nonumber.
\end{eqnarray}
Thus, (\ref{bound}) can be re-written as:
\begin{eqnarray}
\begin{bmatrix}x\\1\end{bmatrix}^T\begin{bmatrix}u_{11}&u_2&\frac{b_1}{2}\\ u_2 & u_{22} & \frac{b_2}{2}\\ \frac{b_1}{2} & \frac{b_2}{2} & r\end{bmatrix}\begin{bmatrix}x\\1\end{bmatrix} = \begin{bmatrix}x\\1\end{bmatrix}^TQ\begin{bmatrix}x\\1\end{bmatrix}&=& 0.\nonumber
\end{eqnarray}
Let $b_1/2 = u_3$ and $b_2/2 = u_4$. There exists an Euclidean transformation, $T_E$, that transforms $Q$ to $\hat{Q}$, such that $\hat{Q}$ is a diagonal matrix and it satisfies:
\begin{eqnarray}
\left[T_E\begin{bmatrix}x\\1\end{bmatrix}\right]^T \hat{Q} \left[T_E\begin{bmatrix}x\\1\end{bmatrix}\right] &=& 0\nonumber.
\end{eqnarray}
Furthermore, there exists an affine transformation (rotation, translation and shear), $T_A$, that transforms $\hat{Q}$ to $\bar{Q}$, where:
\begin{eqnarray}
\label{QBAR}\bar{Q} = \begin{bmatrix}1 & 0 & 0\\ 0 & 1 & 0\\ 0 & 0 & \Lambda\end{bmatrix}; \,\,\,\,\,
T_A\left[T_E\begin{bmatrix}x\\1\end{bmatrix}\right]^T \bar{Q} T_A\left[T_E\begin{bmatrix}x\\1\end{bmatrix}\right] = 0
\end{eqnarray} 
for some $\Lambda\in \R$. Note that the above equation represents a circle, in the new coordinate system, 
where $-\Lambda$ is square of the radius of the circle. These transformations involve pre-multiplying $T_E$ and $T_A$ to $Q$: $\bar{Q} = T_AT_EQ$; transformations are equivalent to some set of elementary row transformations applied to matrix $Q$. There exist many such transformations\footnote{Note that the transformations are nothing but converting $Q$ to a Reduced-Row Echelon form.}. One such sequence of row transformations is:
\begin{eqnarray}
R_3&\leftarrow& R_3-R_1\frac{u_3}{u_{11}}\nonumber; \,\,\,\,\,\,\,
R_2\leftarrow R_2-R_1\frac{u_2}{u_{11}}\nonumber\\
R_3&\leftarrow&R_3-R_2\frac{u_4 - \frac{u_2u_3}{u_{11}}}{S(u_{11})};\nonumber\,\,\,
R_1\leftarrow R_1-R_2\frac{u_2}{S(u_{11})}\nonumber\\
R_2&\leftarrow&R_2-R_3\left(\frac{u_{4}-\frac{u_3u_2}{u_{11}}}{r-\frac{u_3^2}{u_{11}} - (u_{4}-\frac{u_3u_2}{u_{11}})^2S(u_{11})^{-1}}\right)\nonumber
\end{eqnarray}
\begin{eqnarray}
R_1&\leftarrow&R_1-R_3\left(\frac{u_{3} - (u_{4}-\frac{u_3u_2}{u_{11}})u_2 S(u_{11})^{-1}}
{r-\frac{u_3^2}{u_{11}} - (u_{4}-\frac{u_3u_2}{u_{11}})^{2}{S(u_{11})^{-1}}}\right)\nonumber\\
R_1&\leftarrow&\frac{R_1}{u_{11}}\nonumber;\,\,\,\,\, 
R_2\leftarrow\frac{R_2}{S(u_{11})}\nonumber
\end{eqnarray}
Here: $R_i\leftarrow R_i - R_j\beta$ means that the components of $j^{\textit{th}}$ row are multiplied by $\beta$ and then component-wise subtracted from $i^\textit{th}$ row; and $S(u_{11}) = u_{22} - u_2u_{11}^{-1}u_2$ is the Schur Complement of $u_{11}$ in matrix $P$. 
Applying these transformations to $Q$ gives the matrix $\bar{Q}$~(\ref{QBAR}) with following $\Lambda$:
\begin{equation}
\label{lambda}\Lambda = r-\frac{u_3^2}{u_{11}} - {(u_4 - \frac{u_2u_3}{u_{11}})^2}{S(u_{11})}^{-1} = -\mathcal{R}_1^2.
\end{equation}
Matrix $P$ is positive definite if and only if $u_{11}>0$ and $S(u_{11})>0$. Thus, as $P\in S^2_{++}$:
\begin{eqnarray}
\label{use0}
S(u_{11})>0; \,\,\,\,\,\,\text{ and }\,\,\,\,\,\,\,\,\, u_{11}>0.
\end{eqnarray}
In the limiting cases ($\lim_{S(u_{11})\rightarrow0^+}$, and $\lim_{u_{11}\rightarrow0^+}$), it is trivial to show that:
%
%
%
%
\begin{eqnarray}
\lim_{S(u_{11})\rightarrow 0^+} \mathcal{R}_1 = \infty;\nonumber\text{ and } \lim_{u_{11}\rightarrow 0^+}\mathcal{R}_1 = \infty\,\,(\text{as $S(u_{11})>0$}).
\end{eqnarray}
For the limiting case, i.e when $\mathcal{R}_1$ approaches $\infty$, there is nothing to prove. If $\mathcal{R}_1 = \infty$, then it means that there are no surrounding obstacles and the ellipsoid is unbounded. Thus, it remains to show that $\mathcal{R}_1 > 0$, or equivalently: $\Lambda<0$. 
%
%
%
%
By combining (\ref{lambda}) and (\ref{use0}), note that to prove $\Lambda<0$ it remains to show that $r<0$. 
%
%
%

The SDP in RAW constrains the robot to lie inside the ellipsoid. The four corner locations at time $t$, $\Xi(C_t)$,  on the robot's body are constrained to lie inside $\Psi_t$. Thus, the robot's position $(x_t,y_t)$ also lies inside the ellipsoid. Therefore,
\begin{eqnarray}
\begin{bmatrix}x_t\\y_t\end{bmatrix}^TP_t\begin{bmatrix}x_t\\y_t\end{bmatrix} + q_t^T\begin{bmatrix}x_t\\y_t\end{bmatrix} + r_t&\leq&-1\nonumber
\end{eqnarray}
Note that this SDP can be solved in the robot's local coordinate frame, where the robot's current position is the origin and its current navigation direction is the $x-$axis. In fact, RAW solves this SDP in the robot's local coordinate frame for stability. Thus, $(x_t, y_t)$ are both $0, \forall t$ . Substituting this in the above inequality, and removing time-stamp, gives:
\begin{eqnarray}
\begin{bmatrix}0\\0\end{bmatrix}^TP\begin{bmatrix}0\\0\end{bmatrix} + \label{rl1}q^T\begin{bmatrix}0\\0\end{bmatrix} + r\leq-1\implies \,\,\,r\leq-1.
\end{eqnarray}
This completes the proof for (A).
Note that the SDP can also be solved in the global coordinate system, and it will require expanding the square terms and then using a more complicated relation between the variables instead of $r<0$.

To prove (B), applying the identical row transformations for $\{-1\}$ level-set gives:
%
%
%
%
\begin{eqnarray}
\label{eqR2}\mathcal{R}_2 &=& \sqrt{- (r + 1) + \frac{u_3^2}{u_{11}} + \frac{ \left( u_4 - \frac{u_2u_3}{u_{11}}\right)^2 }{S(u_{11})}}.
\end{eqnarray}
By combining (\ref{eqR2}) with (\ref{use0}) and (\ref{rl1}) we get $\mathcal{R}_2\geq0$. Since $\mathcal{R}_1>0$ and $\mathcal{R}_2\geq0$, to show $\mathcal{R}_1-\mathcal{R}_2>0$, it suffices to show $\mathcal{R}_1^2-\mathcal{R}_2^2>0$. We have $\mathcal{R}_1^2 - \mathcal{R}_2^2 = 1$.
This completes the proof. 
\end{proof}

%
%
%
%
\textbf{LEMMA-3:} The new ellipsoid $\Psi_{t+1}$ is formed such that $\Xi(C_{t+1})\in \Gamma_{t}^{t+1}$. The robot thus navigates through a sequence of overlapping ellipsoids.

\begin{proof} From Lemma-1 and 2, we have that $\Xi(C_{t+1})\in \Psi_t$. 
Let $O_{t+1}$ be the point-cloud of obstacles in the robot's current FOV. This point cloud is guaranteed to be separated from $\Xi(C_{t+1})$ with an ellipsoid, otherwise these obstacles would have been discovered in the robot's FOV at time $t$. Also, $\Delta T$ is always selected such that the robot does not reach the waypoint (except when it coincides with the goal). This ensures that the robot never reaches the boundary of the FOV even if the waypoint is generated at the boundary of the FOV. Thus, an obstacle cannot suddenly appear in front of the robot, i.e., at zero-distance from the robot. Hence, there exists an ellipsoid $\Psi_{t+1}$ such that $\Xi(C_{t+1})\in\Psi_{t+1}$ and is separated from $O_{t+1}$. Furthermore, as $\Xi(C_{t+1})\in\Psi_t$, we have that $\Xi(C_{t+1})\in \Gamma_t^{t+1}$.\end{proof}

\textbf{THEOREM:} Let $C_0$ be such that $\Xi(C_0) \in \Psi_0$, i.e., the robot starts from a feasible configuration. Then $\Psi_t$ exists $\forall t=1,...,\infty$, and $\Xi(C_{t})\in\Gamma_{t}^{t+1}$ $\forall t=0,...,\infty$. The robot thus navigates through a sequence of overlapping ellipsoids and is guaranteed to avoid collision throughout the navigation task.

\begin{proof}Follows by combining Lemma-1, 2 and 3.\end{proof}

The complete algorithm is listed in Algorithm 1, and is self-explanatory using the discussions in the previous sections.
\begin{algorithm}[t]                      
\caption{RAW}          
\label{ECAN-Algorithm}                           
\begin{algorithmic}                    
\STATE - initialize parameters: $R_{FOV}, \theta_{FOV}, dr, d\theta, \epsilon, t=0$
\STATE - set a maximum limit for $\Delta T$
\STATE - initialize the policy $\pi$, learned in AWGS architecture
\STATE - initialize robot's position, orientation and goal: $z_0,\theta_0$, $z_g$\\
\STATE
\STATE compute $N$ (number of grid-points in the FOV)
\STATE {\,\,} \textbf{while} ($||z_t-z_g||_2>\epsilon$)
\STATE {} \text{ } \text{ } $O^t\leftarrow$ \textit{getPointCloudObstaclesInFOV}
\STATE {} \text{ } \text{ } \textit{computeFeaturesForEachGridPointInFOV}
\STATE {} \text{ } \text{ } \textit{solveSDP}
\STATE {} \text{ } \text{ } \textbf{for}: $j=\{1,...,N\}$
\STATE {} \text{ } \text{ } \text{ } \text{ } \text{ } \textbf{If} $(\lambda_j\geq 1)$
\STATE {} \text{ } \text{ } \text{ } \text{ } \text{ } \textbf{} \text{} \textbf{} mark $j^{\it th}$ grid-point as infeasible
\STATE {} \text{ } \text{ } \text{ } \text{ } \text{ } \textbf{endif}
\STATE {} \text{ } \text{ } \textbf{endfor}
\STATE {} \text{ } \text{ } \textit{generateWaypointUsingPolicy} $\pi$
\STATE {} \text{ } \text{ } \textit{computeTrajectoryToWaypoint}
\STATE {} \text{ } \text{ } \textit{followTrajectoryForTime} $\Delta T$
\STATE {} \text{ } \text{ } $t\leftarrow t+1$; update: $z_t, \theta_t$
\STATE \textbf{endwhile}
\end{algorithmic}
\end{algorithm}
%
%
%
\section{Experiments: Simulation Results}
\label{Experiments}
This section empirically demonstrates that RAW guarantees safety in unknown environments, cluttered with both the convex and arbitrarily shaped obstacles. To show that RAW generates close to optimal trajectories, RAW trajectories are compared with the global optimal and RRT trajectories --- showing that the actions selected by the RL agent lead to acceptable trajectories. RAW uses the same parameters as in previous work~\cite{AWGS}. These are: 
$\{R_\textit{FOV},dr,\theta_\textit{FOV},d\theta\} =\{5,0.2,60,1^\circ\}$; $\alpha_1=200$ in the reward function. The robot has dimensions $1\times1 m^2$. RAW is implemented in MATLAB, running on a $64$-bit Windows $7$ notebook with core-i7 2.2 GHz and $8$ GB RAM. RRT was run $10$ times for each start-goal configuration, in each of the environmental set-ups. The default maximum RRT iterations is $10^4$ in each run. RAW follows a trajectory to the waypoint for time $\Delta T$, which can be at most $1\, s$, and then a new waypoint is generated.
%
%
%
%
\subsection{Performance Evaluation: Convex Obstacles}
\label{NavConvex}
\begin{figure}
\begin{center}
\includegraphics[trim =10mm 8mm 12mm 1mm, width=6.8cm]{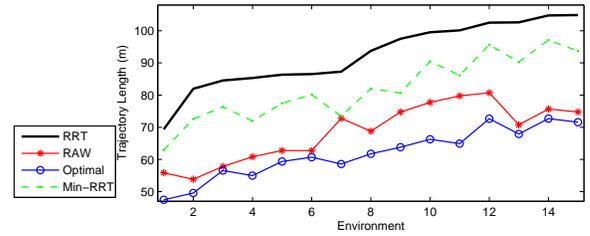}
\end{center}
\captionsetup{font=footnotesize}\vspace{-0.05cm}
\caption{This figure shows that RAW trajectories are shorter than both the average and minimum length (Min-RRT) RRT trajectories in all configurations. RAW trajectories are at most $24\%$ longer than the optimal trajectories. Environments are sorted by the average RRT performance.}
\label{fig:CCD}\vspace{-0.35cm}
\end{figure}

\begin{figure}
\begin{center}
\includegraphics[trim =14mm 8mm 12mm 2mm, width=5.5cm]{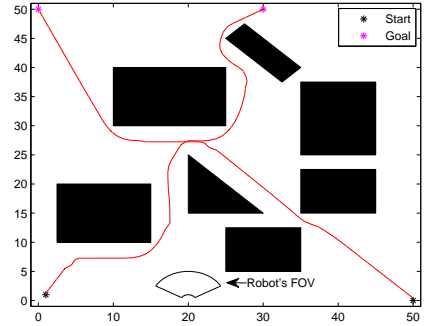}
\end{center}
\captionsetup{font=footnotesize}\vspace{-0.2cm}
\caption{Sample RAW trajectories for two start-goal configurations. The robot's FOV is also shown.}
\label{fig:AAC}\vspace{-0.5cm}
\end{figure}

These experiments empirically show that RAW generates safe trajectories and measure the optimality of RAW trajectories by comparing the trajectory lengths with the optimal and RRT trajectories. The environment has $7$ convex obstacles (shown in Figure~\ref{fig:AAC}). The planners were run in $15$ different start-goal configurations in this environment. RRT trajectories were averaged over $10$ trials for each configuration. Figure~\ref{fig:CCD} compares the algorithms' performance. RAW trajectories are shorter than both the average and minimum length (Min-RRT) RRT trajectories. RAW trajectories are longer than the optimal trajectories. The maximum ratio of the length of RAW and optimal trajectories is $1.24$. Thus, RAW trajectories are at most $24\%$ longer than the optimal trajectories. %
%
%
%
%
%
%
%
Figure~\ref{fig:AAC} shows sample RAW trajectories for two start-goal configurations. It can be seen that the trajectories maintain a safe distance, taking robot's dimensions into account, from the obstacles.
%
%
%
%
\subsection{Performance Evaluation: Arbitrary Obstacles}
\begin{figure}
\begin{center}
\includegraphics[trim =10mm 0mm 10mm 1mm, width=6.5cm]{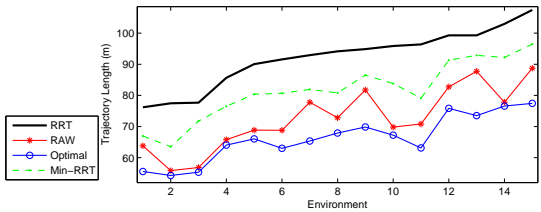}
\end{center}
\captionsetup{font=footnotesize}\vspace{-0.35cm}
\caption{This figure shows that RAW trajectories are shorter than both the average and minimum length RRT (Min-RRT) trajectories. RAW trajectories are at most $19\%$ longer than the optimal trajectories. Environments are sorted by the average RRT performance.}
\label{fig:CCND}\vspace{-0.2cm}
\end{figure}

\begin{figure}
\begin{center}
\includegraphics[trim = 14mm 8mm 12mm 6mm, width = 5.5cm]{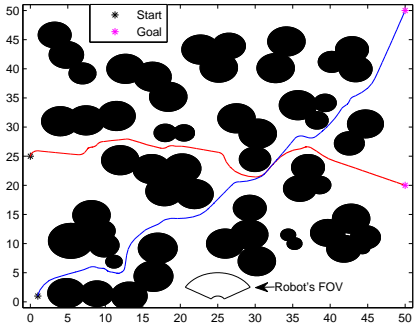}
\end{center}
\captionsetup{font=footnotesize}\vspace{-0.3cm}
\caption{This figure shows sample RAW trajectories for two start-goal configurations shown in blue and red. The robot's FOV is also shown.}
\label{fig:AANC}\vspace{-0.6cm}
\end{figure}

In this experiment the planners are compared in an environment with arbitrary shaped obstacles (Figure~\ref{fig:AANC}), in $15$ different start-goal configurations. Figure~\ref{fig:CCND} shows the numerical results. These environments pose a great challenge for RAW as it has to generate waypoints appropriately to ensure safety of the robot and also minimize the motion time. It can be seen that RAW trajectories are longer than the optimal trajectories. The maximum ratio of the length of RAW and optimal trajectories is $1.19$. Thus, RAW trajectories are at most $19\%$ longer than the optimal trajectories. Furthermore, RAW trajectories are shorter than both the average and minimum length RRT trajectories. Thus RAW generates acceptable trajectories when planning among arbitrarily shaped obstacles in unknown environments. 
%
%
%
%
Figure~\ref{fig:AANC} shows sample RAW trajectories for two start-goal configurations. RAW safely avoided non-convex obstacles, and reached the goal in unknown environments.
%
%
%
%
\label{experiments}
%
%
%
%
\subsection{Performance Evaluation: Corridors}
\begin{figure}
\vspace{0.1cm}
\begin{center}
\includegraphics[trim =8mm 0mm 10mm 2mm, width=7.9cm]{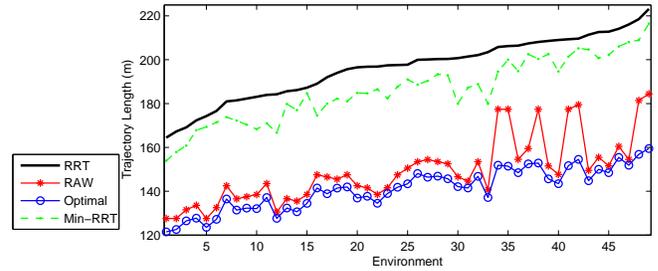}
\end{center}
\captionsetup{font=footnotesize}\vspace{-0.3cm}
\caption{This figures shows that RAW trajectories are shorter than both the minimum length (Min-RRT) and average RRT trajectories. RAW trajectories are longer (at most $17\%$) than the optimal trajectories. Environments are sorted by the average RRT performance.}
\label{fig:CCCD}\vspace{-0.3cm}
\end{figure}

\begin{figure}
\vspace{0.10cm}
\begin{center}
\subfloat[]{\label{fig:Close1}\includegraphics[trim =14mm 10mm 12mm 7mm, width=4.1cm]{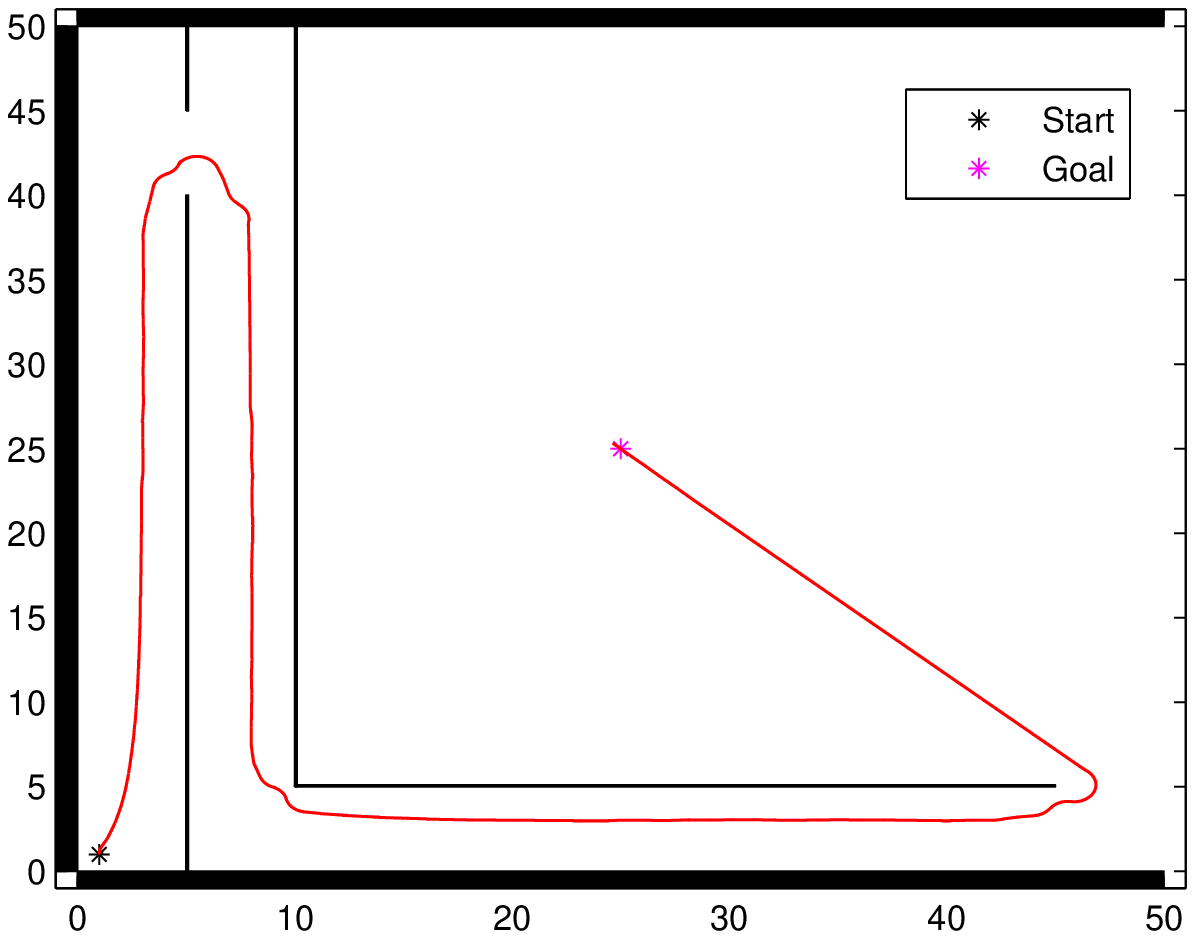}}\,\,\,
\subfloat[]{\label{fig:Close2}\includegraphics[trim =14mm 10mm 12mm 7mm, width=4.1cm]{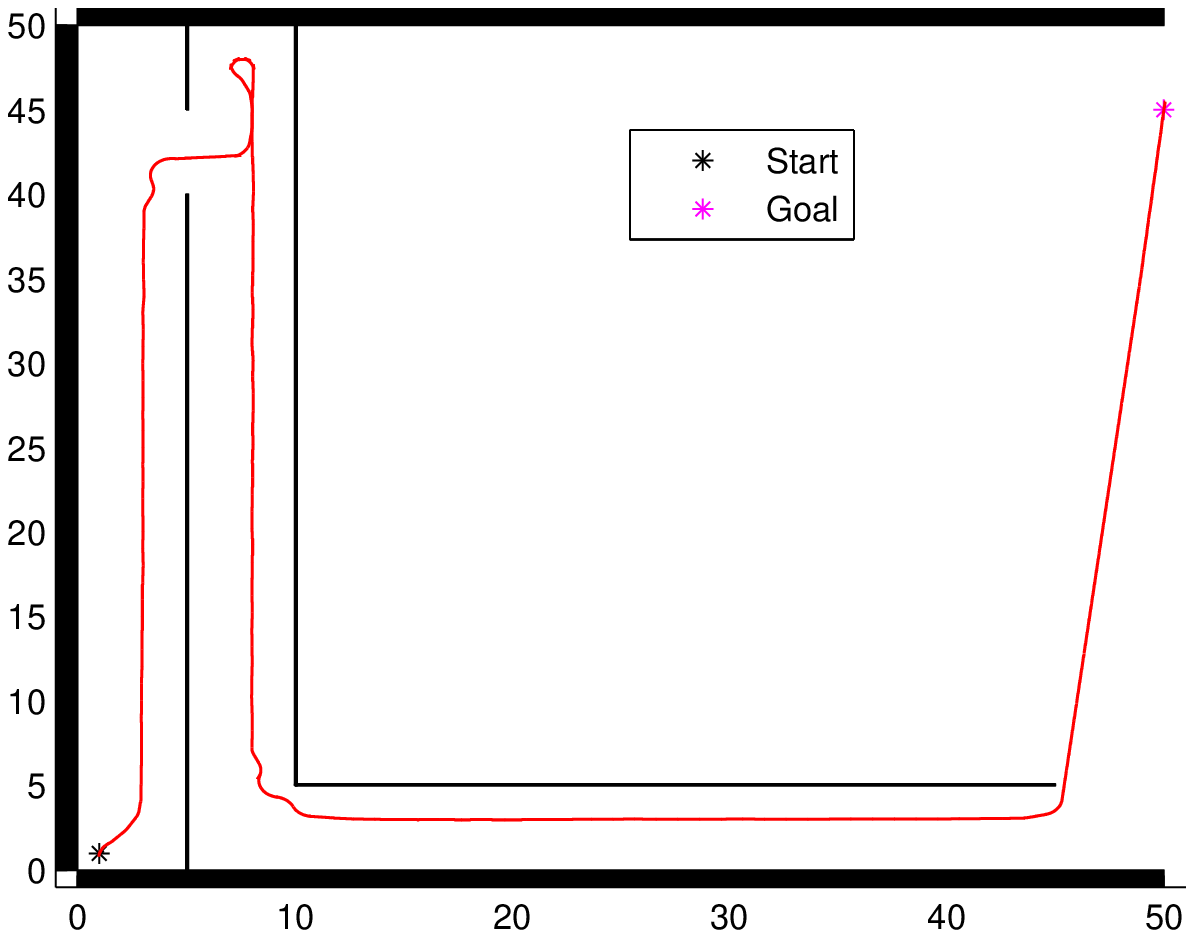}}
\end{center}
\captionsetup{font=footnotesize}\vspace{-0.4cm}
\caption{This figure shows sample RAW trajectories for $2$ start-goal configurations: (a) RAW avoids the dead-ends; and (b) RAW moves towards the blocked region and then turns back to avoid collision in the dead-end.}
\label{fig:Close}\vspace{-0.5cm}
\end{figure}

These experiments empirically show that RAW successfully navigates in an structured environment (as shown in Figure~\ref{fig:Close} --- the environment has corridors, effectively providing a navigation direction) with corridors, in $50$ different start-goal configurations. RRT used $25000$ iterations in each run. 
Figure~\ref{fig:CCCD} shows that both the average and minimum length RRT trajectories are longer than RAW trajectories. As expected, RAW trajectories are longer than the optimal trajectories. The maximum ratio of the length of RAW and optimal trajectories is $1.17$. Thus, RAW trajectories are at most $17\%$ longer than the optimal trajectories. 
Figure~\ref{fig:Close} shows sample RAW trajectories. In Figure~\ref{fig:Close2}, when the goal is at the top-right corner, RAW first visited the blocked region of the second corridor and then turned back to avoid collision. This explains why RAW trajectories are $10-18\,\,m$ (in Figure \ref{fig:CCCD}) longer than the optimal trajectories in $7$ of the $50$ configurations. Also, AWGS collided in the blocked region for the start-goal configuration shown in Figure~\ref{fig:Close2}, as discussed earlier in section-\ref{IC}, while RAW succeeded in avoiding the collision. It should also be noted that one can make the dead-end in Figure~\ref{fig:Close} arbitrarily deep, resulting in possibly longer RAW trajectories. This is because the robot has no prior information about the path being blocked. However, this is true for any planner using limited information, and not just for RAW. 

To show the robustness of RAW for collision avoidance, it is compared with AWGS. AWGS is tested in the same $50$ start-goal configurations. Navigation is considered successful if the robot reaches the goal without colliding with any obstacle. RAW was successful in all $50$ start-goal configurations, while AWGS was successful in only $34$ configurations. RAW successfully avoids collisions by filtering out the potentially dangerous waypoint locations (with SDP) before the collision becomes inevitable, while AWGS lacks such an ability.
%
%
%
%
%
%
%
%
\subsection{Performance Evaluation: Circular Obstacles}
\begin{figure}
\vspace{-0.1cm}
\begin{center}
\includegraphics[trim =14mm 6mm 12mm 0mm, width=7.2cm]{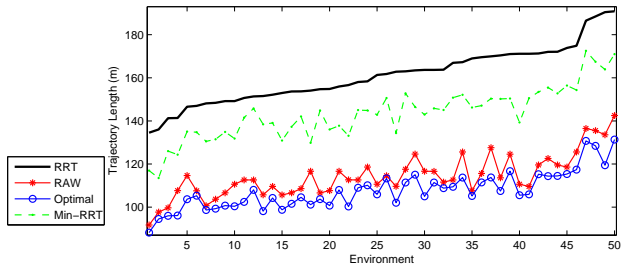}
\end{center}
\captionsetup{font=footnotesize}
\vspace{-0.1cm}
\caption{RAW trajectories are: (i) shorter than both the average and minimum-length (Min-RRT) RRT trajectories and (ii) at most $15\%$ longer than the optimum. 
Environments are sorted by average RRT performance.}
\label{fig:CCOD}\vspace{-0.3cm}
\end{figure}

These experiments measure the optimality of RAW trajectories among $48$ circular obstacles. The planners were tested in $50$ different start-goal configurations. RRT used same parameters as in previous experiment. Figure~\ref{fig:CCOD} shows that RAW trajectories are shorter than both the average and minimum-length RRT trajectories. RAW trajectories are longer than the optimal trajectories. The maximum ratio of the length of RAW and optimal trajectories is $1.15$. Thus, RAW trajectories are at most $15\%$ longer than the optimal trajectories.
%
%
%
%
Figure~\ref{fig:AAOC} shows sample RAW trajectories for two start-goal configurations. 
%
%
%
%
%
%
%
\subsection{Computation Time}
\label{CTime}
\begin{figure}
\vspace{-0.1cm}
\begin{center}
\includegraphics[trim =14mm 8mm 12mm 2mm, width=6.250cm]{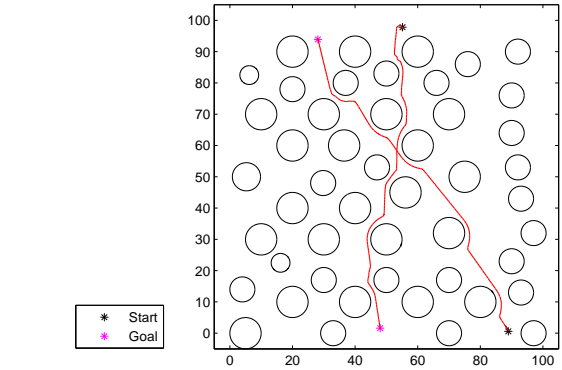}
\end{center}
\captionsetup{font=footnotesize}\vspace{-0.3cm}
\caption{Sample RAW trajectories for two different start-goal configurations.}
\label{fig:AAOC}\vspace{-0.5cm}
\end{figure}

This section discusses the average computation time of RAW --- running in MATLAB on a Core i7, 2.2 GHz notebook computer. The SDP in RAW is solved using the non-commercial solver CVX~\cite{cvx}. RAW was run in $50$ different start-goal configurations of the previous section, in a $105\times105 m^2$ environment shown in Figure~\ref{fig:AAOC}. The average computation time per-step of the algorithm is $98.36\pm8.68 \,\,ms$. Thus, RAW can re-plan at $\approx10$ Hz.
%
%
%
%
\section{Conclusion and Discussion}
\label{Conclusion}
This paper presented an algorithm for on-line reactive motion planning in unknown and unstructured environments that: (i) ensures collision avoidance with unforeseeable obstacles, and (ii) produces trajectories that are not far from the optimal trajectories. RAW was tested in four different experimental setups, testing its performance among all kinds of obstacles. Thus, theoretical claims were verified empirically. Also, RAW succeeded in cases where AWGS failed, showing significant safety improvements over AWGS.

RAW trajectories were at most $15-24\%$ longer than the optimal trajectories. Also, RAW produced shorter trajectories than RRT. This shows that the sequence of locally optimal actions (as taken by the RL agent) generate reasonably acceptable trajectories in unknown environments. However, such a bound ($15-24\%$) cannot be guaranteed in general. For example, as discussed earlier, one can make the dead-end section in Figure~\ref{fig:Close} arbitrarily deep, resulting in longer RAW trajectories. However, this is true for any planner using limited information, and not just for RAW. At least in environments with no-dead ends, similar results are expected. 

Note that, if the environment has dead-ends RAW is not guaranteed to converge to the goal. This is because of the assumption that the robot has no information beyond the FOV. Also, no map is built on-line, and thus the robot may first avoid a dead-end, but then may come back (resulting in an oscillation), when navigating in a corridor type environment. However, this problem of local oscillation is different from the problem of local minima as generally observed in potential field methods. If an on-line map building is allowed, the robot may successfully avoid such oscillations. However, in the potential field methods, at least saddle-points are unavoidable even when the perfect geometric data of the environment is available~\cite{NavigationFunction}.

The robot model assumed in this paper was a simple car-like robot, moving at a constant speed of $1\, m/s$. It was shown that the distance between the level-sets is finite ($\mathcal{R}_1^2-\mathcal{R}_2^2=1$). This ensured safety as the robot can instantaneously change the direction of velocity. For a general robot model, ideally, one would like to have this distance to be equal to the minimum stopping distance of the robot. This would require modifying the SDP and introducing additional constraints. This extension to the SDP was beyond the scope of this paper and is therefore left as a future work.
%
%
%
%
%
%
%
%
\section*{Acknowledgment}
The author gratefully acknowledges the valuable feedbacks from Matthew E. Taylor.

\bibliographystyle{plainnat}
\footnotesize
\bibliography{RAW2}

\end{document}